\pdfoutput=1

\documentclass[11pt]{article}

\usepackage[preprint]{acl}
\usepackage[dvipsnames]{xcolor}
\usepackage{times}
\usepackage{amsmath}
\usepackage{mathtools} 
\usepackage[table]{xcolor}  
\usepackage{color} 
\usepackage{latexsym}
\usepackage{amsmath} 
\usepackage{multirow}
\usepackage{amssymb}
\usepackage[T1]{fontenc}
\usepackage[pdftex]{graphicx} 
\DeclareGraphicsExtensions{.pdf,.png,.jpg,.jpeg} 
\usepackage{caption}
\usepackage{subfigure}
\usepackage{listings}
\usepackage[utf8]{inputenc}
\usepackage{microtype}
\usepackage{tikz}
\usepackage{pgfplots}
\pgfplotsset{compat=1.18} 
\usepackage{xcolor}
\usepackage{arydshln}
\usepackage{array}
\usepackage{booktabs}
\usepackage{inconsolata}
\usepackage{graphicx}
\usepackage{xspace}

\title{Enhancing Language Agent Strategic Reasoning through Self-Play in Adversarial Games}

\newcommand{\method}{\textsc{SCO-PAL}\xspace}

\newcommand{\eg}{\textit{e.g.}\xspace}

\makeatletter
\def\adl@drawiv#1#2#3{%
        \hskip.5\tabcolsep
        \xleaders#3{#2.5\@tempdimb #1{1}#2.5\@tempdimb}%
                #2\z@ plus1fil minus1fil\relax
        \hskip.5\tabcolsep}
\newcommand{\cdashlinelr}[1]{%
  \noalign{\vskip\aboverulesep
           \global\let\@dashdrawstore\adl@draw
           \global\let\adl@draw\adl@drawiv}
  \cdashline{#1}
  \noalign{\global\let\adl@draw\@dashdrawstore
           \vskip\belowrulesep}}
\makeatother

\author{%
 Yikai Zhang$^\spadesuit$
 Ye Rong$^\heartsuit$
 Siyu Yuan$^\heartsuit$
 Jiangjie Chen$^\clubsuit$ 
 Jian Xie $^\diamondsuit$ 
 Yanghua Xiao$^\spadesuit$\thanks{Corresponding author.} \\
\textsuperscript{\rm $\spadesuit$}{Shanghai Key Laboratory of Data Science, School of Computer Science, Fudan University} \quad \\
\textsuperscript{\rm $\clubsuit$}{ByteDance Seed}
\textsuperscript{\rm $\diamondsuit$}{OSU}
\textsuperscript{\rm $\heartsuit$}{School of Data Science, Fudan University} \quad \\
    \texttt{ykzhang22@m.fudan.edu.cn}\quad  \texttt{shawyh@fudan.edu.cn}
}

\begin{document}
\maketitle

\begin{abstract}
Existing language agents often encounter difficulties in dynamic adversarial games due to poor strategic reasoning. 
To mitigate this limitation, a promising approach is to allow agents to learn from game interactions automatically, without relying on costly expert-labeled data.
Unlike static environments where agents receive fixed feedback or rewards, selecting appropriate opponents in dynamic adversarial games can significantly impact learning performance.
However, the discussion of opponents in adversarial environments remains an area under exploration.
In this paper, we propose a \textbf{S}tep-level poli\textbf{C}y \textbf{O}ptimization method through \textbf{P}lay-\textbf{A}nd-\textbf{L}earn, \textbf{\method}. 
Leveraging \method, we conduct a detailed analysis of opponent selection by setting opponents at different levels and find that self-play is the most effective way to improve strategic reasoning in such adversarial environments. 
Utilizing \method with self-play, we increase the average win rate against four opponents by approximately 30\% compared to baselines and achieve a 54.76\% win rate against GPT-4 in six adversarial games. 
\end{abstract}

\section{Introduction}
\label{intro}
Large Language Model (LLM)–based agents have achieved remarkable success in a variety of language-centric tasks—including web navigation~\citep{zhou2023webarena,deng2024mind2web,putta2024agent,iong-etal-2024-openwebagent}, embodied interactions~\citep{wang2023voyager,zhu2023ghost,lin2024swiftsage} and tool invocation~\citep{qin2023toolllm,shen2024hugginggpt}. 
Unlike classic RL agents that operate over a small, well-defined set of discrete or continuous controls and clear state transitions, LLM agents must generate high-dimensional natural-language actions conditioned on long, unstructured textual contexts. 
This fundamentally complicates strategic reasoning in adversarial games, where success hinges on planning over many moves and reacting to an opponent’s evolving strategy under sparse, delayed rewards. 
Benchmarks such as \texttt{GTBench}~\citep{duan2024gtbench} (e.g.\ \texttt{Breakthrough}, \texttt{Connect Four}, \texttt{Nim}) expose these challenges by testing both the depth of planning and the fluency of language-based decision making.

To mitigate the shortcomings in the strategic reasoning of language agents, there are mainly two paths.
One is imitating a strong strategy directly inspired by~\citet{chen2023fireact,zeng2023agenttuning}.
This way is simple and effective, but gathering expert-level data is time-consuming, costly, and challenging to scale.
Therefore, studies on another path focus on \textbf{play-and learn} paradigm, which enriches data through interaction and learning from the interactions.
For example, \citet{song2024trial,xiong2024watch,xi2024agentgym} allow agents to interact autonomously with environments, gathering numerous data for improvement.
However, unlike interacting with unchanging conditions in static environments, the agent in adversarial games faces dynamic and diverse opponents, which brings additional challenges.
In adversarial scenarios, \citet{wang2024sotopia,cheng2024self} enhance the strategic reasoning capabilities of agents with self-play, but they lack analysis of opponent selection in adversarial environments.

Opponent selection has long been recognized as a key factor in adversarial training within the RL community, where methods such as self-play~\citep{silver2017masteringchessshogiselfplay}, opponent sampling~\citep{Arulkumaran2019AlphaStarAE}, and curriculum-based progression~\citep{Bengio2009CurriculumL} are widely used to stabilize learning and encourage generalization.  
However, these techniques are typically designed for structured, low-dimensional action spaces with well-defined state transitions.  
In contrast, LLM-based agents operate in an open-ended, high-dimensional language space, where each ``action'' is a natural language utterance, and each ``state'' is a rich textual history.  
This introduces a qualitatively different set of challenges: opponents not only determine win–lose outcomes, but also shape the semantic distribution and reasoning trajectories the model encounters during learning.
First, if opponents are too strong, their strategy distribution may differ too much from the policy’s, disrupting learning. Playing with weak opponents provides limited positive feedback, slowing policy updates and trapping the model in suboptimal strategies.  
Second, weaker opponents can lead to premature convergence to local optima, limiting further improvement. 
Third, uneven skill levels between agents may also create an imbalanced experience distribution, overemphasizing specific strategies or states while neglecting others.  
Thus, a well-balanced opponent selection strategy is essential for maintaining a diverse training distribution, providing sufficient learning signals, and ensuring a well-rounded experience, ultimately leading to effective training and robust performance.

To better understand how opponents influence agent learning through interaction in adversarial games, a \textbf{play-and-learn} framework is required.  
First, agents need to interact within an adversarial setting.  
These interactions generate numerous actions, where actions in a given state reflect the underlying strategy. 
Second, to refine the strategy, it is crucial to identify which actions are good or bad.  
Finally, an effective method is needed to update the policy and adjust its action distribution accordingly.
Therefore, we propose a new \textbf{S}tep-level poli\textbf{C}y \textbf{O}ptimization method through \textbf{P}lay-\textbf{A}nd-\textbf{L}earn, \textbf{\method}.
In \method, we first assign an opponent to the agent and conduct large-scale game interaction, then estimate step-level rewards using Monte Carlo Estimation. 
Finally, we design a two-stage strategy refinement method using  Kahneman-Tversky Optimization (KTO)~\citep{ethayarajh2024kto} to optimize the strategy.

Using \method, we comprehensively analyze how the opponents affect learning through interaction in strategic environments and find that self-play is the most effective way to conduct the play-and-learn method. 
In experiments, we select six strategic adversarial games from \texttt{GTBench} to evaluate the strategic reasoning capability of \method.
By incorporating \method with self-play, the average win rate against four downstream opponents increases by about 30\% compared to the baselines. 
Notably, the average win rate against GPT-4 achieves 54.76\% across all the games, demonstrating the effectiveness of \method in improving the strategic reasoning of language agents.

In conclusion, our contributions are as follows:
\begin{itemize}
    \item We conduct a comprehensive and detailed analysis of the opponent selection in adversarial environments.

    \item We design a \textbf{S}tep-level poli\textbf{C}y \textbf{O}ptimization method through \textbf{P}lay-\textbf{A}nd-\textbf{L}earn, \textbf{\method}, that significantly optimizes the agent's strategy and improves its performance in adversarial games.
    
    \item We select multiple strategic games and various opponents to evaluate the competitive performance of \method with self-play, finding that the overall win rate increased by about 30\% compared to others, reaching 54.76\% against GPT-4.
\end{itemize}

\section{Related Work}
\paragraph{LLM Agents}
Large language models (LLMs) are increasingly used to power agents for complex tasks~\citep{zhou2023webarena,wang2023voyager,zhu2023ghost,deng2024mind2web,putta2024agent,iong-etal-2024-openwebagent,shen2024hugginggpt}.  
To improve performance, some methods fine-tune LLMs on expert trajectories~\citep{zeng2023agenttuning,chen2023fireact}, while others adopt a play-and-learn paradigm, allowing agents to learn from interactions~\citep{xi2024agentgym,song2024trial,xiong2024watch,wang2024q}.  
These approaches optimize agents via feedback from successful or failed trajectories, using either preference-based or step-wise learning.  
However, most studies assume static environments; we instead focus on adversarial games, where agents must adapt to dynamic, diverse opponents and reason strategically.

\paragraph{Agent in Adversarial Games}
Strategic adversarial games require players to employ strategies, pursuing high-level goals that necessitate reasoning and adapting actions based on opponents' behaviors. 
\citet{huang2024far,costarelli2024gamebench} collect various strategy games to evaluate agents based on LLMs. 
They find that agents still fall significantly short in strategic games, not reaching human-level performance. 
To study the strategic reasoning ability of language agents, \citet{duan2024gtbench} propose \texttt{GTBench}, a language-driven environment designed to evaluate the reasoning capabilities of LLMs in competitive game-theoretic tasks.  
In this work, we use \texttt{GTBench} as a testbed and select several games, such as \texttt{Liar's Dice} and \texttt{Connect4}, for game interaction and evaluation. 
To imitate the limitation of language agents, existing methods~\citep{wang2024sotopia,cheng2024self} train agents by interaction data in gaming environments instead of costly expert-level data.
\citet{wang2024sotopia} proposes to train agents based on the whole successful trajectories, while \citet{cheng2024self} customizes discounted rewards and updates the model through RL.
We design a new play-and-learn method \method, updating the policy through game interaction, step-level reward estimation, and strategy refinement.
Using \method with self-play, we achieve the highest average win rates compared to these methods.

\section{Preliminary}
\subsection{Game Modeling}
The strategic adversarial games from \texttt{GTBench} can be formalized as an episodic Markov Decision Process (MDP), where two players \(\mathcal{P} \in \{\mathcal{P}_{1}, \mathcal{P}_{2}\}\) take turns performing actions in a gaming environment.  
Each game is defined by four elements: the state \(\mathcal{S}\), representing the current game configuration; the observation \(\mathcal{O}\), the visible part of the state for each player; the action space \(\mathcal{A}\); and the transition function \(T: \mathcal{S} \times \mathcal{A} \rightarrow \mathcal{S}\).  
In state \({s}_{i}\), player \(\mathcal{P}_{i}\) observes \({o}_{i}\) and takes an action \({a}_{i} \sim \pi_{i}(\cdot \mid {s}_{i},{o}_{i})\), which transitions the game to state \({s}_{i+1}\).  
The environment returns a reward \({r}\) when a terminal state is reached.  
Each player aims to maximize their expected cumulative reward by optimizing their policy \(\pi_i\).  
In this work, we focus on improving the policy \(\pi_i\) of the language agent \(\mathcal{P}_1\), enabling it to adjust its strategy and achieve higher rewards through interaction.

\subsection{Behavioral Cloning}
Behavioral cloning (BC) is a supervised learning approach where an agent learns to imitate expert behavior by training on expert-level datasets.
Given a dataset \((x, y) \in \mathcal{D}\), the objective of BC is to minimize the following loss:

\vspace{-20pt}
\begin{equation}
L_{BC}(\pi_\theta;\mathcal{D}) = -\mathbb{E}_{(x,y)\sim \mathcal{D}}[\log \pi_\theta(y | x)],
\end{equation}
\vspace{-10pt}

where \(\pi_\theta\) is the model to be updated.

\subsection{Kahneman-Tversky Optimization (KTO)}
Kahneman-Tversky Optimization (KTO)~\cite{ethayarajh2024kto} is an offline reinforcement learning method for preference optimization. 
Unlike Direct Preference Optimization (DPO)~\cite{rafailov2024direct}, KTO requires only an output and a binary reward indicating desirability for a given input, eliminating the need for pair-wise preference data. 
This is advantageous in asymmetric scenarios like werewolf \cite{ye2025multiagentktoreinforcingstrategic}, where creating pair-wise comparisons is challenging. 
Given the datasets \((x, y) \in \mathcal{D}\), KTO optimizes the policy \(\pi_\theta\) using the following loss:

\vspace{-15pt}
\begin{align}
&r_\theta(x,y) = \log \frac{\pi_\theta(y|x)}{\pi_{ref}(y|x)} \\
&z_0 = \mathbb{E}_{(x,y)\sim \mathcal{D}}[KL(\pi_\theta(y|x) \| \pi_{ref}(y|x))] \\
&v(x, y) = \\
& \begin{cases}
\lambda_D \sigma\big(\beta(r_\theta(x, y) - z_0)\big) & \text{if } y \sim y_{\text{desirable}} \mid x \\
\lambda_U \sigma\big(\beta(z_0 - r_\theta(x, y))\big) & \text{if } y \sim y_{\text{undesirable}} \mid x
\end{cases} \\
&L_{KTO}(\pi_\theta;\mathcal{D}) = \mathbb{E}_{(x,y)\sim \mathcal{D}}[\lambda_y - v(x,y)]
\end{align}
\vspace{-15pt}

Here, \( \lambda_D \) and \( \lambda_U \) are hyperparameters representing the losses for desirable and undesirable outcomes, respectively. The parameter \( \lambda_y \) corresponds to \( \lambda_D \) when \( y \) is desirable and \( \lambda_U \) when \( y \) is undesirable.

\section{\method}

\begin{figure*}[t]
    \centering
        \includegraphics[width=1.0\linewidth]{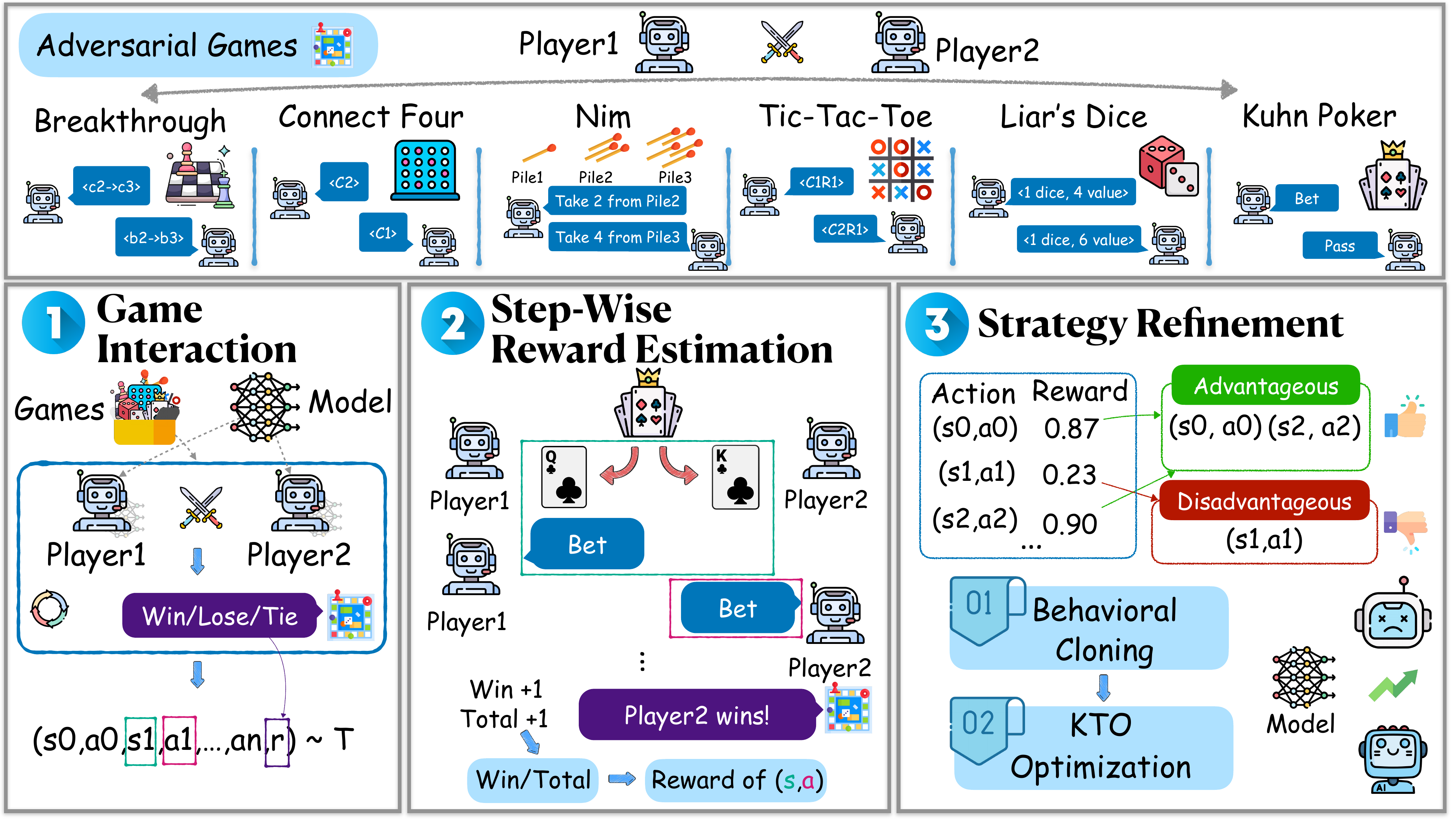} 
    \caption{
    Overview of our play-and-learn method, \method. \method enables language agents to improve through game interaction via a three-stage process:  
Stage \MakeUppercase{\romannumeral 1} Game Interaction assigns an opponent and conducts large-scale gameplay;  
Stage \MakeUppercase{\romannumeral 2} Step-Wise Reward Estimation computes step-level rewards;  
Stage \MakeUppercase{\romannumeral 3} Strategy Refinement updates the policy via behavioral cloning and KTO Optimization.
    }
    \label{fig:framework}
\end{figure*}

To realize learning from interaction, a proper framework is necessary.  
In two-player adversarial games, both players engage in strategic reasoning and act to defeat each other, while the environment autonomously determines the game outcome and allocates rewards based on ending conditions.  
This removes the need for external labeling and facilitates the accumulation of datasets enriched with reward signals.  
Leveraging this, we design a play-and-learn framework \method (Figure~\ref{fig:framework}), using large-scale game interaction to capture strategic optimization directions and improve strategy.

\subsection{Game Interaction}
The play-and-learn paradigm boosts agents by learning from game interaction with opponents.
To realize this, we first conduct interactions between agents in the gaming environments, named Game Interaction (Stage \MakeUppercase{\romannumeral 1}).

During game interaction, player \(\mathcal{P}_{i} \in \{\mathcal{P}_{1}, \mathcal{P}_{2}\}\) performs actions \({a}_t\) according to policy \(\pi_\theta(\cdot \mid {s}_{i},{o}_{i})\) at each time step \(t\).  
Each action \({a}_t\) transitions the game from state \({s}_{t}\) to \({s}_{t+1}\), and the environment concludes the game by determining the ending condition and providing a reward \(r\), indicating a win, loss, or tie.  
For player \(\mathcal{P}_i\), a trajectory \(\tau\) is denoted as \((s_0, a_1, s_1, a_2, \ldots, a_n, s_n, r)\), where \(r\) signifies the outcome reward.  
Through interaction, we collect a set of trajectories \(\mathcal{T} = \{\tau_1, \tau_2, \ldots, \tau_n\}\) for strategy refinement.

\subsection{Step-Wise Reward Estimation}
The environment provides only outcome-level rewards, leaving intermediate actions unlabeled.  
To address this, we design a step-level reward estimation method in Step-Wise Reward Estimation (Stage \MakeUppercase{\romannumeral 2}) and compare different estimation strategies in $\mathsection$~\ref{abl:reward}.  

Step-wise reward estimation identifies whether an action is advantageous or disadvantageous, guiding subsequent strategy refinement.  
We apply Monte Carlo Estimation over interaction trajectories \(\mathcal{T}\), estimating the reward \(r(s_i, a_i)\) of each \((s_i, a_i) \in \tau\) by its empirical win rate.  
Specifically, we count total occurrences \(N_{all}\) and winning occurrences \(N_{win}\) of \((s_i, a_i)\) in \(\mathcal{T}\), and compute:
$r(s_i, a_i) = \frac{N_{win}}{N_{all}}.$

\subsection{Strategy Refinement}
After reward estimation, we obtain trajectories with step-level rewards for training.  
In Strategy Refinement (Stage \MakeUppercase{\romannumeral 3}), we propose a two-stage strategy improvement method:  
(1) \textbf{Behavioral Cloning} adapts the agent to the game environment using state-action pairs \((s_i, a_i)\) with rewards \(r(s_i, a_i)\) above a predefined threshold \(\delta\) as advantageous actions for training.  
(2) \textbf{KTO Optimization} formulates learning as preference optimization.  
We label state-action pairs as advantageous (desirable) or disadvantageous (undesirable) based on \(r(s_i, a_i)\) and \(\delta\), encouraging the agent to prefer advantageous actions while avoiding disadvantageous ones.

In experiments, we find that Behavioral Cloning enables fast adaptation, while KTO Optimization helps the agent effectively distinguish between action qualities, significantly improving win rates in downstream competitions.

\section{Analysis on Opponent Selection}
\label{sec:analysis}

\begin{figure*}[t]
    \centering
    \subfigure[Win rates in game interaction and evaluation.] {
        \pgfplotsset{width=0.35\linewidth, height=0.1\linewidth, compat=1.5, scale only axis} 
\begin{tikzpicture}
    \begin{axis}[
        ylabel={Win Rates},
        ylabel style={font=\footnotesize},
        xmin=0, xmax=18,
        ymin=0, ymax=65,
        xtick={2,4,6,8,10,12,14,16},   
        xticklabels={Random,Self Play, MCTS(5), MCTS(10), MCTS(100), MCTS(200), MCTS(500), MCTS(1000)},
        xticklabel style={rotate=45, font=\tiny},
        ytick={10,30,50}, 
        legend style={fill opacity=0.7, font=\tiny, at={(0.25,0.62)}, anchor=north, legend columns=1}, 
        ymajorgrids=true,
        grid style=dashed
    ]

    \addplot[
        color=NavyBlue,
        mark=triangle,
        ]
        coordinates {
        (2,56.59)(4,49.73)(6,35.24)(8,26.98)(10,7.52)(12,6.47)(14,5.81)(16,5.40)
        };
    \addlegendentry{Game Interaction}

    \addplot[
        color=Red!70,
        mark=o,
        ]
        coordinates {
        (2,44.53)(4,50.08)(6,47.02)(8,41.50)(10,42.64)(12,41.48)(14,40.75)(16,38.74)
        };
    \addlegendentry{Evaluation}

    \end{axis}
\end{tikzpicture}
        \label{fig:dis1_opponent}
    }
    \subfigure[Number of actions collected from different opponents.] {
        \pgfplotsset{width=0.4\linewidth,height=0.1\linewidth,compat=1.5,scale only axis}
\begin{tikzpicture}
    \begin{axis}[
        ylabel={Actions},
        ylabel style={font=\footnotesize},
        ymin=0, ymax=25000,
        symbolic x coords={Random,Self Play,MCTS(5),MCTS(10),MCTS(100),MCTS(200),MCTS(500),MCTS(1000)},
        xtick=data,
        xticklabel style={rotate=45, font=\tiny},
        ytick={10000,20000}, 
        ymajorgrids=true,
        grid style=dashed,
        bar width=10,
        enlarge x limits=0.1,
        legend style={
            fill opacity=0.7,
            draw=black,  
            font=\tiny,
            at={(0.8,0.97)},  
            anchor=north,  
            legend columns=1, 
            /tikz/column 2/.style={column sep=1pt} 
        },
        legend image code/.code={%
            \draw[fill] (0cm,-0.1cm) rectangle (0.16cm,0.1cm); 
        }
    ]
    \addplot[
        ybar,
        fill=red!50,
        ]
        coordinates {
        (Random,17068)
        (Self Play,23906)
        (MCTS(5),14999)
        (MCTS(10),14773)
        (MCTS(100),13084)
        (MCTS(200),11793)
        (MCTS(500),11176)
        (MCTS(1000),10218)
        };
    \addlegendentry{Disadvantageous}

    \addplot[
        ybar,
        fill=NavyBlue,
        ]
        coordinates {
        (Random,9337)
        (Self Play,12210)
        (MCTS(5),5448)
        (MCTS(10),4051)
        (MCTS(100),733)
        (MCTS(200),372)
        (MCTS(500),225)
        (MCTS(1000),78)
        };
    \addlegendentry{Advantageous}

    \end{axis}
\end{tikzpicture}
        \label{fig:dis1_opponent_num}
    }
    \caption{
    Analysis of opponent selection in game interaction.  
    (a) \textit{Game Interaction} shows the base model's win rates against different opponents, while \textit{Evaluation} shows the win rates of the model trained with \method against four downstream opponents.  
    (b) displays the number of actions collected when interacting with different opponents.  
    The number in MCTS(\(\cdot\)) indicates the \texttt{max\_simulation} setting.
    }
    \label{fig:select_opponent}
\end{figure*}
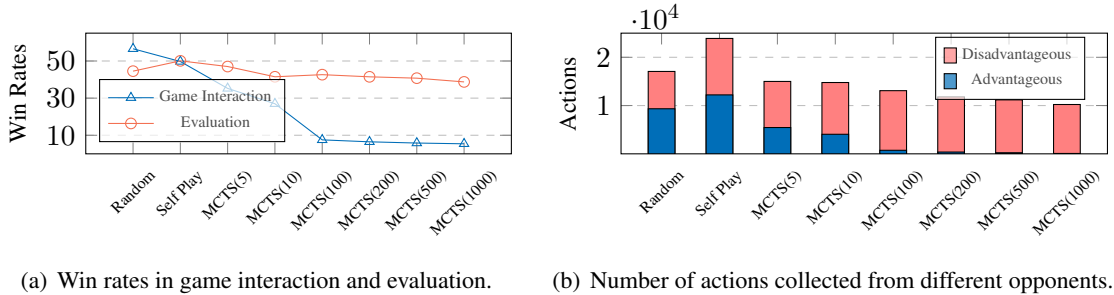

\subsection{Setup}

\paragraph{Game Interaction Opponents}
To study how opponent strength affects learning, we introduce level-controllable symbolic opponents ranging from random (weak) to MCTS-based (strong).  
The strength of MCTS is adjusted by varying the maximum number of simulations (\texttt{max\_simulation})\footnote{Details in Appendix~\ref{app:mcts}.}, set to 5, 10, 100, 200, 500, and 1000.

\paragraph{Game Benchmark}
We select 6 games from \texttt{GTBench}~\cite{duan2024gtbench}: (1) \texttt{Breakthrough}, (2) \texttt{Connect Four}, (3) \texttt{Tic-Tac-Toe}, (4) \texttt{Kuhn Poker}, (5) \texttt{Liar’s Dice}, and (6) \texttt{Nim}\footnote{Game introductions are in Appendix~\ref{app:game_intro}.}.  
Each game interaction involves 1,000 episodes, and each evaluation is conducted over 100 episodes per opponent.  
To mitigate first-player advantage, players alternate roles and play an equal number of games.

\paragraph{Model Setting}  
We use Qwen2-7B-Instruct~\cite{qwen2} as the base model. During game interaction, the temperature is set to 0.7 (see $\mathsection$~\ref{app:temp} for ablation).  
For evaluation, we follow \texttt{GTBench}~\cite{duan2024gtbench} and set the temperature to 0.2. 
We also demonstrate the effectiveness of \method on other models (\eg, Mistral) in the Appendix~\ref{app:mistral}.

\paragraph{Parameters}  
The threshold for identifying advantageous actions is 0.5.  
In Behavioral Cloning, we set the learning rate to 1e-6, batch size to 2, warm-up ratio to 0.1, and train for 5 epochs.  
For KTO, the batch size is 2, the gradient accumulation steps are 8, the learning rate is 1e-6, and training also lasts 5 epochs.  
The loss weights $\lambda_D$ and $\lambda_U$ are set such that $\lambda_D n_D = \lambda_U n_U$, where $n_D$ and $n_U$ are the numbers of advantageous and disadvantageous samples; the larger weight is set to 1.0, and the smaller scaled proportionally.  
All experiments are conducted on 8 NVIDIA A100 GPUs (80GB each).

\paragraph{Metrics}  
We evaluate performance using win rate. For each game, we record outcome counts \(N_{win}\), \(N_{lose}\), and \(N_{tie}\), and compute the win rate as:  
$w = \frac{N_{win} + 0.5N_{tie}}{N_{win}+N_{lose}+N_{tie}}.$

\paragraph{Evaluation Opponents}  
We evaluate against four opponents:  
(1) \textit{Random}, selecting actions at random;  
(2) \textit{MCTS}, configured with 1 rollout, UCT parameter \(c = 2\), and max simulations = 1000;  
(3) \textit{GPT-3.5} (GPT-3.5-turbo-0125)~\cite{openai2024gpt4technicalreport}, an OpenAI language model;  
(4) \textit{GPT-4} (GPT-4-turbo-2024-04-09)~\cite{openai2024gpt4technicalreport}, a more advanced OpenAI model.

\subsection{Results}

\label{exp:choose_oppo}

\paragraph{Opponents affect the number and proportion of actions.}  
We first examine win rates against different opponents.  
As shown in Figure~\ref{fig:dis1_opponent}, the language agent achieves a 56.59\% win rate against the weak opponent (random) and nearly 50\% in self-play.  
Against strong opponents (MCTS), as \texttt{max\_simulation} increases, MCTS yields more accurate state estimations, and the agent’s win rate drops from 35.24\% to 5.40\%, consistent with the expected strength levels.

We then analyze the number of actions collected during the interaction.  
As shown in Figure~\ref{fig:dis1_opponent_num}, the number of actions decreases when facing either very weak or very strong opponents.  
In both cases, the skill gap makes outcomes more predictable, with the advantaged player repeatedly winning through fixed strategies, thereby reducing the diversity of actions.  
In contrast, self-play yields the most balanced ratio of advantageous to disadvantageous actions.  
This ratio becomes increasingly skewed against stronger opponents, where the agent struggles to generate advantageous actions, resulting in a significant imbalance.

\begin{figure}[t]
    \centering
    \subfigure[Number of actions.]{
        \label{fig:exp_scale1}
        \pgfplotsset{width=0.33\linewidth,height=0.2\linewidth,compat=1.5,scale only axis}
\begin{tikzpicture}
    \begin{axis}[
        ylabel={Actions},
        ylabel style={font=\footnotesize},
        ymin=0, ymax=30000,
        symbolic x coords={Self Play,MCTS(5),Scale 1,Scale 2},
        xtick=data,
        xticklabel style={rotate=45, font=\tiny},
        ytick={10000,20000},
        ymajorgrids=true,
        grid style=dashed,
        bar width=10,
        enlarge x limits=0.2,
        legend style={
            draw=black,  
            fill=white,  
            fill opacity=0.75,
            font=\tiny,
            at={(0.5,0.95)},  
            anchor=north,  
            legend columns=1, 
            /tikz/column 2/.style={column sep=1pt} 
        },
        legend image code/.code={%
            \draw[fill] (0cm,-0.1cm) rectangle (0.16cm,0.1cm); 
        }
    ]

    \addplot[
        ybar,
        fill=red!50,
        ]
        coordinates {
        (Self Play,23906)
        (MCTS(5),14999)
        (Scale 1,24600)
        (Scale 2,24611)
        };
    \addlegendentry{Disadvantageous}

    \addplot[
        ybar,
        fill=NavyBlue,
        ]
        coordinates {
        (Self Play,12210)
        (MCTS(5),5448)
        (Scale 1,8382)
        (Scale 2,12589)
        };
    \addlegendentry{Advantageous}
    \end{axis}
\end{tikzpicture}
    }
    \hspace{-0.5cm} 
    \subfigure[Win rates after training.]{
        \label{fig:exp_scale2}
        \pgfplotsset{width=0.33\linewidth,height=0.2\linewidth,compat=1.5,scale only axis}
\begin{tikzpicture}
    \begin{axis}[
        ylabel={Win Rates},
        ylabel style={font=\footnotesize},
        xmin=0, xmax=10,
        ymin=37, ymax=52,
        xtick={2,4,6,8},   
        xticklabels={Self Play,MCTS(5),Scale 1,Scale 2},
        ytick={40,45,50}, 
        ymajorgrids=true,
        grid style=dashed,
        xticklabel style={rotate=45, font=\tiny}
    ]

    \addplot[
        color=NavyBlue,
        mark=triangle,
        ]
        coordinates {
        (2,50.08)(4,47.02)(6,44.73)(8,41.35)
        };

    \end{axis}
\end{tikzpicture}
    }
    \caption{
    Evaluation of different scaling strategies for balancing advantageous and disadvantageous samples.  
    (a) shows the number of actions obtained under each setting. (b) reports the win rates against four downstream opponents using agents trained on the corresponding data.
    }
    \label{fig:scale}
\end{figure}
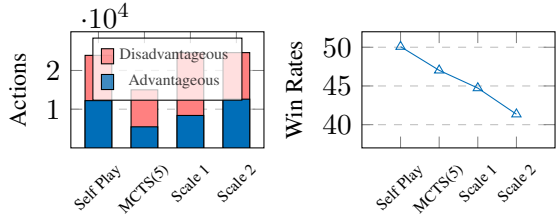

\paragraph{Self-play is the most effective choice.}  
We evaluate \method by training it with different opponents and reporting the average win rate against four downstream opponents.  
As shown in Figure~\ref{fig:dis1_opponent}, the results mirror the action count trends in Figure~\ref{fig:dis1_opponent_num}, with self-play yielding the largest performance improvement, outperforming both weak and strong opponents.  
As opponent strength increases, the diversity and proportion of advantageous actions decline, leading to reduced learning effectiveness.  
These findings underscore the importance of collecting diverse gameplay data and maintaining a balanced distribution of advantageous and disadvantageous actions for effective strategy optimization.

\paragraph{Scaling data does not improve performance.}

Figure~\ref{fig:dis1_opponent_num} shows that self-play yields the richest and most balanced data, leading to more accurate strategy optimization.  
To examine the effect of data volume, we upsample~\footnote{For a given dataset D1, we randomly sample data from D1 to construct a new dataset D2 of the expected upsampling size.} training data under different opponent settings, as shown in Figure~\ref{fig:exp_scale1}.  
Specifically, we consider two scaling strategies:  
(1) \textbf{Scale 1} aligns the total number of samples with that of self-play while preserving the original advantageous-to-disadvantageous ratio.  
(2) \textbf{Scale 2} directly matches the number of advantageous and disadvantageous samples to those from self-play, thereby eliminating action-ratio differences induced by opponent strength.  
In both settings, we use actions collected from MCTS(5) interactions in Figure~\ref{fig:dis1_opponent_num} as the base data.

We train agents on the scaled data and compare performance with self-play and unscaled MCTS(5) in Figure~\ref{fig:dis1_opponent}, with results summarized in Figure~\ref{fig:exp_scale2}.  
We observe that Scale 1 leads to a noticeable performance drop, suggesting that simply increasing sample size without improving action quality can cause overfitting and misguide optimization.  
Likewise, Scale 2 results in a further 3\% decline in win rate, indicating that forcibly balancing the number of samples disrupts the original optimization trajectory.  
These findings suggest that data scaling alone cannot compensate for the lack of high-quality or diverse actions—balance and strategic richness must be intrinsically present in the data.

\paragraph{Potential of combining strong-opponent imitation and \method.}

In our opponent selection analysis, strong opponents play a key role.  
To further explore this, we examine the combination of \method with imitation learning from strong opponents.  
Results in Appendix~\ref{app:immi} and~\ref{app:immi_self} show that imitation can complement \method, offering further performance improvements.

\section{Experiments on \method with Self-Play}
\label{sec:experiment}
Analysis in $\mathsection$~\ref{sec:analysis} demonstrates that self-play is the most effective way to conduct play-and-learn in adversarial games.
To fully explore the performance of \method with self-play, we compare \method under self-play with existing play-and-learn methods, both against downstream opponents, and conduct a head-to-head evaluation in this section.

\newcolumntype{q}{>{\columncolor{green!4}}c}
\newcolumntype{b}{>{\columncolor{Blue!4}}c}
\newcolumntype{p}{>{\columncolor{Purple!10}}c}

\setlength\tabcolsep{1.5pt}

\begin{table*}[t]
\centering
\small
\begin{tabular}{cqbbbbbbp}
\toprule
\textbf{Methods} & \textbf{Opponents} & \textbf{Breakthrough} & \textbf{Connect Four} & \textbf{Tic-Tac-Toe} & \textbf{Kuhn Poker} & \textbf{Liar’s Dice} & \textbf{Nim}  & \textbf{Avg.} \\
\midrule

\multirow{5}{*}{\textbf{Base}}
& GPT-4 & 7.22 & 46.46 & 44.50 & 20.90 & 20.31 & 31.00 & 28.40 \\ 
& GPT-3.5 & 57.65 & 46.32 & 55.26 & 18.75 & 21.54 & 53.00 & 42.09 \\
& Random & 66.67 & 61.62 & 63.64 & 39.06 & 39.66 & 44.00 & 52.44 \\
& MCTS & 2.74 & 0.00 & 6.00 & 20.90 & 4.84 & 0.00 & 5.75 \\
\cdashlinelr{2-9}
& AVG & 33.57 & \underline{38.60} & 42.35 & 24.90 & 21.59 & 32.00 & 32.17 \\
\midrule

\multirow{5}{*}{\textbf{BC}}
& GPT-4 & 26.00 & 32.50 & 64.00 & 45.00 & 57.00 & 31.00 & 42.58 \\ 
& GPT-3.5 & 52.63 & 10.10 & 100.00 & 45.00 & 68.69 & 44.68 & 53.52 \\
& Random & 68.09 & 66.00 & 78.00 & 61.00 & 83.00 & 50.00 & 67.68 \\
& MCTS & 3.30 & 0.00 & 0.00 & 45.00 & 13.00 & 0.00 & 10.22 \\
\cdashlinelr{2-9}
& AVG & \underline{37.50} & 27.15 & \textbf{60.50} & \textbf{49.00} & \underline{55.42} & 31.42 & \underline{43.50} \\
\midrule

\multirow{5}{*}{\textbf{SPAG}}
& GPT-4 & 29.87 & 42.00 & 33.00 & 41.00 & 57.00 & 39.00 & 40.31 \\ 
& GPT-3.5 & 30.14 & 29.29 & 98.00 & 37.00 & 64.65 & 44.00 & 50.51 \\
& Random & 71.43 & 65.00 & 75.00 & 45.45 & 82.00 & 57.00 & 65.98 \\
& MCTS & 0.00 & 0.00 & 0.00 & 45.00 & 15.00 & 0.00 & 10.00 \\
\cdashlinelr{2-9}
& AVG & 32.86 & 34.07 & 51.50 & 42.11 & 54.66 & \underline{35.00} & 41.70 \\
\midrule

\multirow{5}{*}{\textbf{\method}}
& GPT-4 & 52.53 & 62.63 & 71.43 & 45.00 & 60.00 & 37.00 & 54.76 \\ 
& GPT-3.5 & 60.42 & 72.00 & 54.64 & 45.00 & 80.00 & 77.00 & 64.84 \\
& Random & 85.88 & 76.00 & 76.02 & 35.00 & 89.00 & 59.00 & 70.15 \\
& MCTS & 2.33 & 0.00 & 10.10 & 45.00 & 6.00 & 0.00 & 10.57 \\
\cdashlinelr{2-9}
& AVG & \textbf{50.29} & \textbf{52.66} & \underline{53.05} & \underline{42.50} & \textbf{58.75} & \textbf{43.25} & \textbf{50.08} \\

\bottomrule
\end{tabular}
\caption{Win rates against four opponents on six games. The best results are \textbf{bolded}, and the second best ones are \underline{underlined}. Statistical significance tests are reported in the Appendix~\ref{app:signi}.}
\label{tab:main}
\end{table*}

\subsection{Setup}

\paragraph{Opponents}
In game interaction, we set the opponent as the language agent itself (self-play setting in $\mathsection$~\ref{sec:analysis}).
Other settings are the same as these in $\mathsection$~\ref{sec:analysis}.

\paragraph{Baselines}
We select three baselines: (1) Base model: we use Qwen2-7b-Chat as the base model; (2) Behavioral Cloning (BC): fine-tuning the model on successful trajectories from game interactions inspired by~\citet{wang2024sotopia}; (3) SPAG~\cite{cheng2024self}: that updates the policy by hand-crafted rewards with a discounted coefficient~\footnote{Detailed introduction of baselines are in Appendix~\ref{app:baseline}.}.

\subsection{Main Results}
\label{exp:main_results}
In Table~\ref{tab:main}, the base model's average win rate against the four downstream opponents is generally lower, only slightly beating random with a rate of 52.44\%. 
After training, \method achieves the most significant improvement, 50.08\%, surpassing 43.50\% of BC and 41.70\% of SPAG, and defeating GPT-4 with an average winning rate of 54.76\%.
The reason behind the improvement is that both BC and SPAG struggle with learning actions due to inaccurate reward estimation and insufficient optimization methods. 
For example, in BC, successful trajectories might include suboptimal actions. 
In SPAG, manually specified coefficients can become inaccurate and not general due to changes in environments. 
\method uses more general reward estimates and finer-grained reward signals, using KTO optimization, leading to more comprehensive learning.
We also calculate regret, with results in the Appendix~\ref{app:regret} showing that \method reduces regret and improves strategic reasoning.

\subsection{Head-to-Head Evaluation}
\paragraph{Settings}
In this experiment, we use the six games from \texttt{GTBench} to conduct direct battles between two players, with 100 games played in each match.

\newcolumntype{k}{>{\columncolor{purple!4}}c}
\newcolumntype{j}{>{\columncolor{pink!4}}c}
\newcolumntype{u}{>{\columncolor{blue!4}}c}

\setlength\tabcolsep{5pt}

\begin{table}[t]
\small 
  \centering
    \begin{tabular}{kjjjju}
    \toprule
      & \textbf{Base} & \textbf{BC} & \textbf{SPAG} & \textbf{\method} & \textbf{Avg.} \\ 
    \midrule
    \textbf{Base} & {\color[HTML]{A9A9A9} 50.00} & 27.20 & 24.21 & 22.97 & 31.10 \\ 
    \textbf{BC} & 72.80 & {\color[HTML]{A9A9A9} 50.00} & 51.03 & 45.29 & 54.78 \\
    \textbf{SPAG} & 75.79 & 48.97 & {\color[HTML]{A9A9A9} 50.00} & 47.31 & 55.52 \\ 
    \textbf{\method} & \textbf{77.03} & \textbf{54.71} & \textbf{52.69} & {\color[HTML]{A9A9A9} 50.00} & \textbf{58.61} \\ 
    \bottomrule
    \end{tabular}
    \caption{The vertical axis represents Player 1, the horizontal axis represents Player 2, and the values indicate the win rate of Player 1 against Player 2.}
  \label{tab:headtohead}
\end{table}

\paragraph{Results}
As shown in Table~\ref{tab:headtohead}, \method achieves a win rate of over 50\% compared to all the others. 
While BC and SPAG can also outperform the base model, both of them struggle against our \method. 
This indicates that our approach effectively identifies the advantageous and disadvantageous actions and uses step-level strategy optimization to achieve a consistently high win rate.

\subsection{Generalization on Unseen Games}
\paragraph{Settings}
In $\mathsection$~\ref{exp:main_results}, \method demonstrates effectiveness on in-domain games, showing its improvement in the language agent's strategic reasoning ability. 
To explore the generalization potential of \method, we further select three unseen games, \texttt{Blind
Auction}, \texttt{Pig} and \texttt{Prisoner’s Dilemma}~\footnote{Introductions of unseen games are in appendix~\ref{app:unseen}.} from \texttt{GTbench} and test the performance of \method against a random strategy.

\paragraph{Results}
As shown in Table~\ref{tab:dis5_ood}, the strategic reasoning ability developed by \method also transfers to unseen games, with an average improvement of 8\% across three games. 
In these games, we find that the agent can make reasonable decisions and timely strategy adjustments based on the current state and winning goals, even being unseen before, indicating that \method enhances the language agent's strategic reasoning capability~\footnote{Experiments on general capabilities are in Appendix~\ref{app:general}.}.

\subsection{Iteration of \method with Self-Play}
We analyze the performance of applying \method with self-play iteratively and find that as iterations increase, the agent's performance initially improves but later declines. 
This demonstrates that our \method supports effective iteration but carries a slight risk of overfitting due to RL training. See the Appendix~\ref{app:iteration} for detailed experiments.

\section{Ablations}
\subsection{Ablation of Training Strategies in \method}

To validate the effectiveness of our two-stage approach, we conduct ablations on both the training procedure and the data used for BC.

\paragraph{Ablation of Training Strategy}  
We compare three training strategies:  
(1) \textbf{Direct KTO}, applying KTO without BC;  
(2) \textbf{Joint Loss}, combining BC and KTO objectives;  
(3) \textbf{Two-Stage (Ours)}, performing BC followed by KTO.  
As shown in Table~\ref{fig:ab1_sft}, adding BC improves adaptation to the game environment, with the two-stage strategy achieving a 4.59\% performance gain over the joint-loss variant.  
These results suggest that initializing with BC provides a more stable foundation for subsequent preference optimization.

\begin{table}[t]
\small 
  \centering
    \begin{tabular}{kjjju}

\toprule
 & \textbf{Auc.} & \textbf{Pig} & \textbf{Pri.} & \textbf{Avg.} \\  
\midrule
\textbf{Base} & \textbf{71.00} & 88.00 & 37.00 & 65.00  \\ 
\textbf{\method} & 67.00 & \textbf{92.00} & \textbf{60.00} & \textbf{73.00}  \\ 
\bottomrule

\end{tabular}
    \caption{Evaluation on unseen games. \textbf{Auc.} refers to \texttt{Blind Auction}; \textbf{Pig} refers to \texttt{Pig}; \textbf{Pri.} refers to \texttt{Prisoner’s Dilemma}.} 
\label{tab:dis5_ood}
\end{table}

\paragraph{Ablation of Data for Behavioral Cloning}  
In \method, BC is trained on advantageous actions identified from game interaction.  
We compare two BC data construction methods based on self-play:  
(1) \textbf{Trajectory-Based}, which uses all actions from successful trajectories;  
(2) \textbf{Reward-Based}, which filters actions with estimated rewards exceeding a threshold.  
As shown in Table~\ref{fig:ab2_sftway}, using only advantageous actions yields better results, as successful trajectories may still contain suboptimal decisions.  
This highlights the importance of fine-grained filtering in BC data selection.

\paragraph{Ablation of Strategy Optimization}  
We compare strategy optimization methods within \method and find that adding KTO on top of BC significantly improves performance by reinforcing advantageous actions and suppressing disadvantageous ones, as shown in Table~\ref{fig:ab3_po}.  
This highlights the importance of strategy optimization after self-play.
We further compare KTO with DPO and observe a 9.56\% gain in average win rate.  
Unlike DPO, which relies on pairwise comparisons under identical states, KTO operates at the step level and offers broader coverage of action quality, leading to more effective optimization.

\begin{figure}[t]
\centering
\subfigure[Training Strategy]{
  \label{fig:ab1_sft}
  \pgfplotsset{width=0.32\linewidth,height=0.2\linewidth,compat=1.18,scale only axis}
\begin{tikzpicture}
    \begin{axis}[
        ybar,
        nodes near coords,
        xtick=\empty,  
        ylabel={Win Rates},
        bar width=10pt,
        ymin=32, ymax=52,
        axis y line*=left,
        font=\footnotesize,
        axis x line*=bottom,
        enlarge x limits=0.8,
        ymajorgrids=true,
        grid style=dashed,
        legend style={
            draw=black,  
            fill=white,  
            font=\tiny,
            at={(0.52,0.02)},  
            anchor=south,  
            legend columns=3, 
            /tikz/column 2/.style={column sep=1pt} 
        },
        legend image code/.code={%
            \draw[fill] (0cm,-0.1cm) rectangle (0.16cm,0.1cm); 
        }
    ]
       
    \addplot[color=Blue, fill=Blue!30] coordinates {(1.25, 44.89)};
    \addlegendentry{Direct}  

    \addplot[color=BlueGreen, fill=BlueGreen!30] coordinates {(1.3, 45.49)};
    \addlegendentry{Joint}  

    \addplot[color=Lavender, fill=Lavender!50] coordinates {(1.35, 50.08)};
    \addlegendentry{Ours}  
    \end{axis}
\end{tikzpicture}
} \hspace{-0.2cm}
\subfigure[Data of BC]{
  \label{fig:ab2_sftway}
  \pgfplotsset{width=0.32\linewidth,height=0.2\linewidth,compat=1.18,scale only axis}
\begin{tikzpicture}
    \begin{axis}[
        ybar,
        nodes near coords,
        xtick=\empty,  
        bar width=10pt,
        ymin=32, ymax=52,
        axis y line*=left,
        font=\footnotesize,
        axis x line*=bottom,
        enlarge x limits=0.8,
        ymajorgrids=true,
        grid style=dashed,
        legend style={
            draw=black,  
            fill=white,  
            font=\tiny,
            at={(0.52,0.02)},  
            anchor=south,  
            legend columns=1, 
            /tikz/column 2/.style={column sep=1pt} 
        },
        legend image code/.code={%
            \draw[fill] (0cm,-0.1cm) rectangle (0.16cm,0.1cm); 
        }
    ]
        \addplot[color=Blue, fill=Blue!30] coordinates {(1.2, 43.50)};
    \addlegendentry{Traj-based}  

    \addplot[color=BlueGreen, fill=BlueGreen!30] coordinates {(1.4, 46.87)};
    \addlegendentry{Reward-based}  
    \end{axis}
\end{tikzpicture}
} \\
\subfigure[Preference Optimization]{
  \label{fig:ab3_po}
  \pgfplotsset{width=0.32\linewidth,height=0.2\linewidth,compat=1.18,scale only axis}
\begin{tikzpicture}
    \begin{axis}[
        ybar,
        nodes near coords,
        xtick=\empty,  
        ylabel={Win Rates},
        bar width=10pt,
        ymin=32, ymax=52,
        axis y line*=left,
        font=\footnotesize,
        axis x line*=bottom,
        enlarge x limits=0.8,
        ymajorgrids=true,
        grid style=dashed,
        legend style={
            draw=black,  
            fill=white,  
            font=\tiny,
            at={(0.52,0.02)},  
            anchor=south,  
            legend columns=3, 
            /tikz/column 2/.style={column sep=1pt} 
        },
        legend image code/.code={%
            \draw[fill] (0cm,-0.1cm) rectangle (0.16cm,0.1cm); 
        }
    ]
    \addplot[color=Blue, fill=Blue!30]  coordinates {(1.25, 46.31)};
    \addlegendentry{BC}  

    \addplot[color=BlueGreen, fill=BlueGreen!30] coordinates {(1.3, 50.08)};
    \addlegendentry{+KTO}  

    \addplot[color=Lavender, fill=Lavender!50]  coordinates {(1.35, 40.52)};
    \addlegendentry{+DPO}  
    \end{axis}
\end{tikzpicture}
} \hspace{-0.2cm}
\subfigure[Reward Estimation]{
  \label{fig:ab4_reward}
  \pgfplotsset{width=0.32\linewidth,height=0.2\linewidth,compat=1.18,scale only axis}
\begin{tikzpicture}
    \begin{axis}[
        ybar,
        nodes near coords,
        xtick=\empty,  
        bar width=10pt,
        ymin=32, ymax=52,
        axis y line*=left,
        font=\footnotesize,
        axis x line*=bottom,
        enlarge x limits=0.8,
        ymajorgrids=true,
        grid style=dashed,
        legend style={
            draw=black,  
            fill=white,  
            font=\tiny,
            at={(0.35,0.02)},  
            anchor=south,  
            legend columns=1, 
            /tikz/column 2/.style={column sep=1pt} 
        },
        legend image code/.code={%
            \draw[fill] (0cm,-0.1cm) rectangle (0.16cm,0.1cm); 
        }
    ]
    \addplot[color=Blue, fill=Blue!30]  coordinates {(1.25, 48.66)};
    \addlegendentry{Discounted}  

    \addplot[color=BlueGreen, fill=BlueGreen!30] coordinates {(1.3, 50.08)};
    \addlegendentry{Win Rate}  

    \addplot[color=Lavender, fill=Lavender!50]  coordinates {(1.35, 36.77)};
    \addlegendentry{Beta}  
    \end{axis}
\end{tikzpicture}
}
\caption{Ablations of \method.}
\label{fig:abl}
\end{figure}

\subsection{Ablation of Reward Estimation Methods}
\label{abl:reward}

Different reward estimation methods produce varying distributions of advantageous and disadvantageous actions, thus influencing \method's performance.  
We compare three approaches—\textbf{Discounted Reward}, \textbf{Win Rate}, and \textbf{Beta}—as detailed in Appendix~\ref{app:reward}.
As shown in Figure~\ref{fig:ab4_reward}, the win rate method performs best with a 50.08\% win rate, compared to 48.66\% for discounted reward and 36.77\% for beta.  
Win rate directly captures success likelihood, aligning well with the game objective and offering a stable, interpretable signal.  
In contrast, discounted reward requires careful tuning of the discount factor, while the beta method depends on sufficient data and sensitive hyperparameters, reducing practicality.  
Overall, win rate offers the best trade-off between simplicity, stability, and alignment, making it the most effective choice in \method.

\section{Conclusion}
In this paper, we focus on improving the strategic reasoning ability of language agents under the play-and-learn paradigm.
We design a \textbf{S}tep-level poli\textbf{C}y \textbf{O}ptimization method through \textbf{P}lay-\textbf{A}nd-\textbf{L}earn, \method.
Using \method, we analyze how opponents influence agent learning in adversarial games and find that self-play yields the best results.
Incorporating \method with self-play, we boost the
average win rate by 30\% compared to baselines and achieve a 54.76\% win rate against GPT-4. 
Our work provides insights into opponent selection in adversarial environments and introduces a new learning framework, offering an effective way to enhance the strategic reasoning of language agents.

\section*{Limitations}
While \method demonstrates strong empirical performance in symbolic adversarial games, it has certain limitations.  
First, our study is limited to well-defined, turn-based games and does not explore more open-ended or partially observable environments such as negotiation or web-based tasks.  
Second, the opponent pool is restricted to scripted agents and LLM variants; incorporating human or style-diverse LLM opponents could further enrich strategic diversity but introduces additional challenges.  
We leave these directions for future work.

\section*{Ethical Statement}
This work focuses on improving the strategic reasoning abilities of language agents in adversarial games using self-play and interaction-based learning.  
All experiments are conducted in controlled, simulated environments without real user data or deployment.  
Our use of existing artifacts is consistent with their intended use, and we follow their license and terms.
No sensitive data or human annotations are used.  
While advances in strategic language modeling may enable more capable agents in competitive settings, we caution against misuse in domains such as manipulation, deception, or adversarial dialogue generation.  
We encourage future research to incorporate safety constraints and ethical safeguards when deploying such agents in open environments.

\bibliography{acl_latex}

\appendix


\label{sec:appendix}
\clearpage
\appendix
\section*{\Large{Appendix}}

\section{Supplementary Experiments}

\subsection{Imitating Strong Players}
\label{app:immi}

In $\mathsection$~\ref{sec:analysis}, we examine how opponents influence performance by introducing stronger ones. Strong opponents tend to adopt better strategies, leading to higher win rates. To further leverage their potential, we explore the effect of imitating strong players.

\begin{figure}[t]
    \centering
    \subfigure[Number of actions.] {
        \pgfplotsset{width=0.8\linewidth,height=0.3\linewidth,compat=1.5,scale only axis}
\begin{tikzpicture}
    \begin{axis}[
        ylabel={Actions},
        ylabel style={font=\footnotesize},
        ymin=0, ymax=65000,
        symbolic x coords={MCTS(5),MCTS(10),MCTS(100),MCTS(200),MCTS(500),MCTS(1000)},
        xtick=data,
        xticklabel style={rotate=45, font=\footnotesize},
        ytick={20000,40000,60000}, 
        ymajorgrids=true,
        grid style=dashed,
        bar width=10,
        enlarge x limits=0.1,
        legend style={
            draw=black,  
            fill opacity=0.75,
            font=\footnotesize,
            at={(0.7,0.97)},  
            anchor=north,  
            legend columns=1, 
            /tikz/column 2/.style={column sep=1pt} 
        },
        legend image code/.code={%
            \draw[fill] (0cm,-0.1cm) rectangle (0.16cm,0.1cm); 
        }
    ]

    \addplot[
        ybar,
        fill=red!50,
        ]
        coordinates {
        (MCTS(5),46518)
        (MCTS(10),44986)
        (MCTS(100),57442)
        (MCTS(200),57305)
        (MCTS(500),57119)
        (MCTS(1000),56748)
        };
    \addlegendentry{Disadvantageous}

    \addplot[
        ybar,
        fill=NavyBlue,
        ]
        coordinates {
        (MCTS(5),23905)
        (MCTS(10),22748)
        (MCTS(100),27631)
        (MCTS(200),27833)
        (MCTS(500),28098)
        (MCTS(1000),28103)
        };
    \addlegendentry{Advantageous}

    \end{axis}
\end{tikzpicture}
        \label{fig:dis2_immi_num}
    }
    \hspace{-0.5cm}
    \subfigure[Win rates: relative and evaluation.] {
        \pgfplotsset{width=0.8\linewidth,height=0.3\linewidth,compat=1.5,scale only axis}
\begin{tikzpicture}
    \begin{axis}[
        ylabel={Win Rates},
        ylabel style={font=\footnotesize},
        xmin=0, xmax=14,
        ymin=0, ymax=65,
        xtick={2,4,6,8,10,12},   
        xticklabels={MCTS(5),MCTS(10),MCTS(100),MCTS(200),MCTS(500),MCTS(1000)},
        xticklabel style={rotate=45, font=\footnotesize},
        ytick={25,50}, 
        ymajorgrids=true,
        grid style=dashed,
        legend style={
            draw=black,  
            fill=white,  
            font=\footnotesize,
            at={(0.7,0.35)},  
            anchor=south,  
            legend columns=1, 
        },
        legend cell align={left},
    ]

    \addplot[
        color=NavyBlue,
        mark=triangle,
        ]
        coordinates {
        (2,54.58)(4,61.19)(6,57.02)(8,52.57)(10,51.79)(12,49.92)
        };
    \addlegendentry{Evaluation}  

    \addplot[
        color=red!30,
        mark=o,
        ]
        coordinates {
        (2,35.24)(4,26.98)(6,7.52)(8,6.47)(10,5.81)(12,5.40)
        };
    \addlegendentry{Relative}
    \end{axis}
\end{tikzpicture}
        \label{fig:dis2_immitation}
    }
    \caption{(a) shows the number of actions taken during self-play between strong players of varying levels. In (b), \textit{Relative} denotes the win rate of the base model against different opponents, while \textit{Evaluation} indicates the average win rate of the model trained by imitating strong players, evaluated against four downstream opponents. The number in MCTS(·) specifies the \texttt{max\_simulation} parameter.}
    \label{fig:immi}
\end{figure}
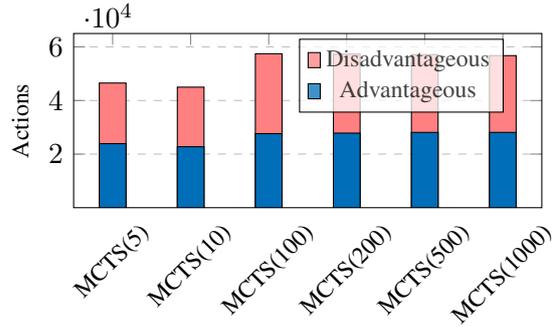
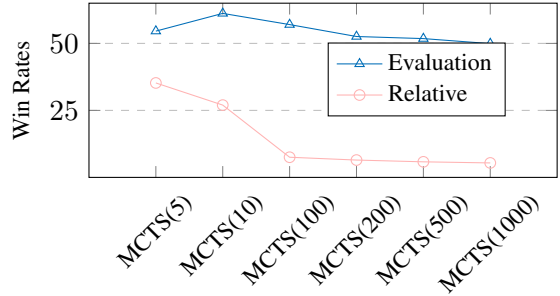

We first experiment with direct imitation. As shown in $\mathsection$~\ref{sec:analysis}, self-play generates diverse, balanced, and high-quality data through interactive gameplay. To better capture strong players’ strategies, we adapt \method accordingly. Two strong agents of equal strength play against each other, followed by step-wise reward estimation. We then select advantageous samples with win rates above a threshold for BC training. As shown in Figure~\ref{fig:immi}, learning from strong players significantly improves win rates.

Moreover, we observe that as the opponent’s relative strength (compared to the base model) increases, the agent’s performance initially improves but later declines. This indicates that while imitating strong players is beneficial, a large skill gap and divergent strategies can impede effective learning.

\subsection{Combining \method and Imitating Strong Players}
\label{app:immi_self}

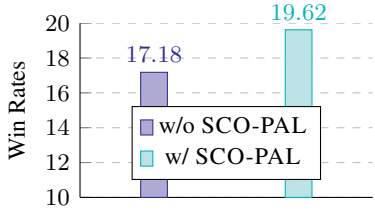
\begin{figure}[t] 
    \centering
    \pgfplotsset{width=0.5\linewidth,height=0.3\linewidth,compat=1.5,scale only axis}
\begin{tikzpicture}
    \begin{axis}[
        ybar,
        nodes near coords,
        xtick=\empty,  
        ylabel={Win Rates},
        bar width=10pt,
        ymin=10, ymax=20,
        axis y line*=left,
        font=\footnotesize,
        axis x line*=bottom,
        enlarge x limits=0.8,
        ymajorgrids=true,
        grid style=dashed,
        legend style={
            draw=black,  
            fill=white,  
            font=\footnotesize,
            at={(0.5,0.1)},  
            anchor=south,  
            legend columns=1, 
            /tikz/column 2/.style={column sep=1pt} 
        },
        legend image code/.code={%
            \draw[fill] (0cm,-0.1cm) rectangle (0.16cm,0.1cm); 
        }
    ]
    \addplot[color=Blue, fill=Blue!30] coordinates {(1.2, 17.18)};
    \addlegendentry{w/o \method}  

    \addplot[color=BlueGreen, fill=BlueGreen!30] coordinates {(1.4, 19.62)};
    \addlegendentry{w/ \method}  
    \end{axis}
\end{tikzpicture} 
    \caption{Win rate changes against MCTS(1000) after applying \method{} with self-play to the model trained by imitating MCTS(5).}
    \label{fig:immi_selfplay}
\end{figure}

In $\mathsection$\ref{app:immi}, we observed that learning from strong players enhances strategic reasoning. Here, we take the best-performing setting—imitating MCTS(5)—and apply \method with self-play. We evaluate the resulting model against a strong opponent, MCTS(1000). As shown in Figure~\ref{fig:immi_selfplay}, \method remains effective even after expert imitation, improving the average win rate against strong opponents by 2.44\%. This suggests that imitating strong players can complement \method with self-play, leading to further gains.

\subsection{Discussion on Iteration}
\label{app:iteration}

\paragraph{Settings}
During iteration, the agent plays against historical models in a play-and-learn setup. In the first round, as there is no historical model, the base model (Qwen2-7B-Chat) performs self-play, and we apply \method to obtain \textit{Iter1}. In the second round, \textit{Iter1} plays against the base model, is trained to yield \textit{Iter2}, and then \textit{Iter2} plays against \textit{Iter1} to produce \textit{Iter3}.

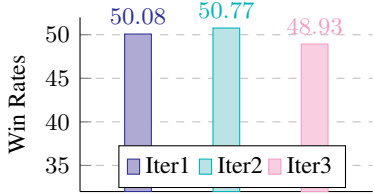
\begin{figure}[t]  
    \centering
    \pgfplotsset{width=0.5\linewidth,height=0.3\linewidth,compat=1.5,scale only axis}
\begin{tikzpicture}
    \begin{axis}[
        ybar,
        nodes near coords,
        xtick=\empty,  
        ylabel={Win Rates},
        bar width=10pt,
        ymin=32, ymax=52,
        axis y line*=left,
        font=\footnotesize,
        axis x line*=bottom,
        enlarge x limits=0.8,
        ymajorgrids=true,
        grid style=dashed,
        legend style={
            draw=black,  
            fill=white,  
            font=\footnotesize,
            at={(0.52,0.02)},  
            anchor=south,  
            legend columns=3, 
            /tikz/column 2/.style={column sep=1pt} 
        },
        legend image code/.code={%
            \draw[fill] (0cm,-0.1cm) rectangle (0.16cm,0.1cm); 
        }
    ]
    \addplot[color=Blue, fill=Blue!30]  coordinates {(1.25, 50.08)};
    \addlegendentry{Iter1}  

    \addplot[color=BlueGreen, fill=BlueGreen!30] coordinates {(1.3, 50.77)};
    \addlegendentry{Iter2}  

    \addplot[color=Lavender, fill=Lavender!50]  coordinates {(1.35, 48.93)};
    \addlegendentry{Iter3}  
    \end{axis}
\end{tikzpicture} 
    \caption{Performance across iterations of \method. Each iteration involves self-play and training against previous models.}
    \label{fig:iter}
\end{figure}

\paragraph{Results}
Figure~\ref{fig:iter} presents the results of iteratively applying \method. We observe that iterative self-play with \method gradually improves the model’s strategic performance. However, in \textit{Iter3}, the win rate declines slightly. We attribute this to overfitting introduced by repeated RL training~\cite{zeng2023agenttuning,xiong2024watch}, which may narrow the model’s strategic diversity and reduce effectiveness.

\subsection{Impact of Temperature}
\label{app:temp}

Temperature is a key parameter in LLM generation, influencing the diversity of outputs. To assess its effect on \method, we evaluate five settings: 0.2, 0.5, 0.7, 1.0, and 1.2, and systematically analyze how temperature impacts performance.

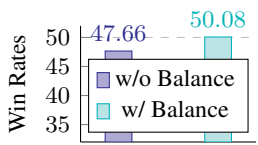
\begin{figure}[b] 
    \centering
    \pgfplotsset{width=0.3\linewidth,height=0.2\linewidth,compat=1.5,scale only axis}
\begin{tikzpicture}
    \begin{axis}[
        ybar,
        nodes near coords,
        xtick=\empty,  
        ylabel={Win Rates},
        bar width=10pt,
        ymin=32, ymax=52,
        axis y line*=left,
        font=\footnotesize,
        axis x line*=bottom,
        enlarge x limits=0.8,
        ymajorgrids=true,
        grid style=dashed,
        legend style={
            draw=black,  
            fill=white,  
            font=\footnotesize,
            at={(0.5,0.08)},  
            anchor=south,  
            legend columns=1, 
            /tikz/column 2/.style={column sep=1pt} 
        },
        legend image code/.code={%
            \draw[fill] (0cm,-0.1cm) rectangle (0.16cm,0.1cm); 
        }
    ]
    \addplot[color=Blue, fill=Blue!30] coordinates {(1.2, 47.66)};
    \addlegendentry{w/o Balance}  

    \addplot[color=BlueGreen, fill=BlueGreen!30] coordinates {(1.4, 50.08)};
    \addlegendentry{w/ Balance}  
    \end{axis}
\end{tikzpicture} 
    \caption{Comparison of win rates with and without the balance mechanism.}
    \label{fig:balance}
\end{figure}

\paragraph{Temperature affects action diversity and game validity.}

\begin{figure*}[t]
    \centering
	\subfigure[Number of actions] {
	\label{fig:ab5_temp1}
		\pgfplotsset{width=0.2\linewidth,height=0.15\linewidth,compat=1.5,scale only axis}
\begin{tikzpicture}
    \begin{axis}[
        ylabel={Actions},
        xmin=0.1, xmax=1.3,
        ymin=8000, ymax=26000,
        xtick={0.2,0.5,0.7,1.0,1.2},   
        ytick={10000, 15000, 20000, 25000}, 
        ymajorgrids=true,
        grid style=dashed,
    ]

    \addplot[
        color=blue!50,
        mark=triangle,
        ]
        coordinates {
        (0.2,9110)(0.5,20128)(0.7,23906)(1.0,22336)(1.2,13729)
        };

    \end{axis}
\end{tikzpicture}}
	\subfigure[Game validity rate] {
    \label{fig:ab5_temp2}
		\pgfplotsset{width=0.2\linewidth,height=0.15\linewidth,compat=1.5,scale only axis}
\begin{tikzpicture}
    \begin{axis}[
        ylabel={Valid Rates},
        xmin=0.1, xmax=1.3,
        ymin=20, ymax=80,
        xtick={0.2,0.5,0.7,1.0,1.2},   
        ytick={25, 50, 75}, 
        ymajorgrids=true,
        grid style=dashed,
        scaled y ticks=false
    ]
    \addplot[
        color=purple!50,
        mark=triangle,
        ]
        coordinates {
        (0.2,67.85)(0.5,67.75)(0.7,65.6)(1.0,57.68)(1.2,41.62)
        };

    \end{axis}
\end{tikzpicture}}
    \subfigure[Post-training win rate] {
    \label{fig:ab5_temp3}
		\pgfplotsset{width=0.2\linewidth,height=0.15\linewidth,compat=1.5,scale only axis}
\begin{tikzpicture}
    \begin{axis}[
        ylabel={Win Rates},
        xmin=0.1, xmax=1.3,
        ymin=42, ymax=52,
        xtick={0.2,0.5,0.7,1.0,1.2},   
        ytick={40, 45, 50}, 
        ymajorgrids=true,
        grid style=dashed,
    ]
    \addplot[
        color=green!30,
        mark=triangle,
        ]
        coordinates {
        (0.2,42.93)(0.5,47.03)(0.7,50.08)(1.0,44.80)(1.2,46.76)
        };

    \end{axis}
\end{tikzpicture}}
    \caption{Effect of temperature on (a) number of actions during interaction, (b) validity rate of generated games, and (c) average win rate after training with data from each temperature setting.}
    \label{fig:task_input}
\end{figure*}
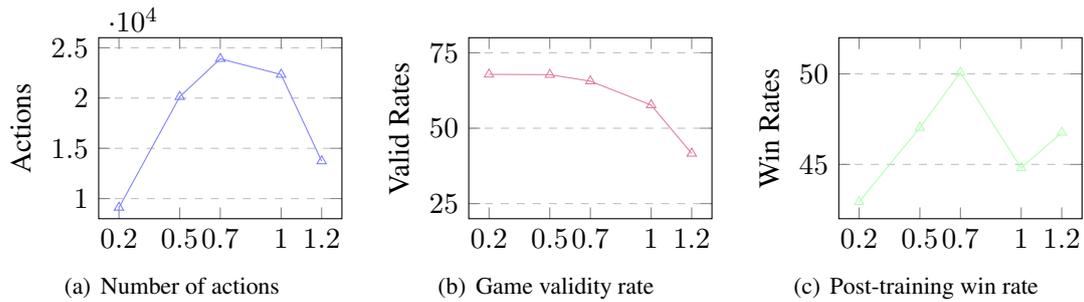

Temperature controls the token sampling behavior of language models. Lower temperatures lead to more deterministic outputs, while higher values introduce greater diversity. In our setting, temperature influences the diversity of actions during game interaction by the language agent, which in turn affects the number of actions taken.

As shown in Figure~\ref{fig:ab5_temp1}, action diversity increases with temperature up to a point, peaking between 0.7 and 1.0, then declines. To understand this trend, we examine the rate of valid games—those where the agent’s actions are successfully parsed by the environment. At high temperatures, the agent is more likely to generate invalid actions, reducing the number of valid interactions. Figure~\ref{fig:ab5_temp2} shows that the valid rate peaks between 0.5 and 0.7.

Based on this analysis, we set the temperature to 0.7 during self-play. We also evaluate \method under different temperature settings, as shown in Figure~\ref{fig:ab5_temp3}. Results confirm that a temperature of 0.7 yields the highest post-training win rate, indicating that a balance between stability and diversity of valid actions is crucial for optimal performance.

\subsection{Impact of Game Balance}

Different games have varying action spaces, leading to an unequal number of actions after self-play. To avoid biased learning, we apply data balancing by upsampling actions from games with fewer actions and downsampling those with more, while keeping the total sample size constant. We compare results before and after balancing. As shown in Figure~\ref{fig:balance}, this strategy ensures the language agent learns effectively across different games.

\subsection{Statistical Significance Tests}
\label{app:signi}
\begin{table}[t]
\small 
  \centering
    
    \vspace{0.1cm}
    \begin{tabular}{kj}
\toprule
 & \textbf{P-Value} \\  
\midrule
\textbf{BC} & 0.0017  \\ 
\textbf{SPAG} & 0.0110  \\ 
\bottomrule

\end{tabular}
\caption{Significance test for the main experiment.} 
\label{tab:significant}
\end{table}
We perform significance tests comparing our method against baselines (BC and SPAG). As shown in Table~\ref{tab:significant}, all $p$-values are below 0.05, indicating that the improvements are statistically significant.

\subsection{Generalization Capability Evaluation}
\label{app:general}
\begin{table}[t]
\small 
  \centering
    \begin{tabular}{kj}
\toprule
 & \textbf{Accuracy.} \\  
\midrule
\textbf{Base} & 70.54  \\ 
\textbf{\method} & 70.65  \\ 
\bottomrule

\end{tabular}
\caption{Model performance on a general benchmark (MMLU).} 
\label{tab:dis_general}
\end{table}

We assess whether applying \method affects the model’s generalization. Specifically, we evaluate performance on MMLU~\cite{hendrycks2021measuringmassivemultitasklanguage} before and after training. As shown in Table~\ref{tab:dis_general}, \method yields a slight improvement over the base model. This suggests some generalization to unseen tasks, although the gain remains limited due to the breadth and diversity of the benchmark.

\subsection{\method on Other Models}
\label{app:mistral}
\begin{table}[b]
\small 
  \centering
    \begin{tabular}{kj}
\toprule
 & \textbf{Average Win Rates.} \\  
\midrule
\textbf{Base} & 32.18  \\ 
\textbf{\method} & 45.80  \\ 
\bottomrule
\end{tabular}
\caption{\method on Mistral.} 
\label{tab:mistral}
\end{table}

To test the transferability of \method, we apply it to Mistral-7B-Instruct-v0.3~\cite{jiang2023mistral7b}. As shown in Table~\ref{tab:mistral}, \method also improves strategic reasoning in Mistral, demonstrating its versatility across different base models.

\subsection{Regret Analysis}
\label{app:regret}
\begin{table}[t]
\small 
  \centering
    \begin{tabular}{kjj}
\toprule
 & \textbf{Nim} & \textbf{Tic-Tac-Toe} \\  
\midrule
\textbf{Base} & 0.0564 & 0.0658  \\ 
\textbf{\method} & \textbf{0.0378} & \textbf{0.0472}   \\ 
\bottomrule
\end{tabular}
\caption{Regret on Nim and Tic-Tac-Toe.} 
\label{tab:regret}
\end{table}

To assess progress toward equilibrium, we measure regret in two benchmark games—Nim and Tic-Tac-Toe—before and after applying \method. Results in Table~\ref{tab:regret} show consistent regret reduction, indicating enhanced strategic alignment.

\onecolumn
\section{Method Settings}

\subsection{MCTS Search Algorithm}
\label{app:mcts}
Monte Carlo Tree Search (MCTS)~\cite{chaslot2008monte} is a heuristic search algorithm used for decision-making processes, particularly in game playing. 
The core principle of MCTS is to build a search tree incrementally and asymmetrically by using random simulations to explore the most promising moves. The algorithm consists of four main steps: selection, expansion, simulation, and backpropagation.
\begin{itemize}
    \item \textbf{Selection:} Starting from the root node, select child nodes based on a policy that balances exploration and exploitation. The most common policy is the Upper Confidence Bound for Trees (UCT), defined as:
    \[
    UCT = \frac{w_i}{n_i} + c \times \sqrt{\frac{\ln N}{n_i}}
    \]
    where \( w_i \) is the number of wins for the node, \( n_i \) is the number of times the node has been visited, \( N \) is the total number of simulations, and \( c \) is the exploration constant.
    \item \textbf{Expansion:} Once a leaf node is reached, expand the tree by adding one or more child nodes corresponding to possible moves. In our setup, the tree expands with each new node representing possible future states.
    \item \textbf{Simulation:} From the new node, perform a rollout, i.e., play out the game randomly until a terminal state is reached. The outcome of this simulation provides an estimate of the node's value.
    \item \textbf{Backpropagation:} Use the result of the simulation to update the information in the nodes along the path from the expanded node back to the root. This involves updating win ratios and visit counts to reflect the newly gathered information.
\end{itemize}

In our settings, we set \texttt{rollout\_count}: The number of rollouts performed from each new node is set to 1; \texttt{c}: The exploration constant of UCT is set to 2, balancing the trade-off between exploration of new nodes and exploitation of known promising nodes; \texttt{max\_simulations}: The maximum number of simulations. This parameter controls the total number of simulations that can be performed during decision-making processes.
By adjusting \texttt{max\_simulations}, we can control the level of agents, where a higher number of simulations generally leads to more informed and strategic decisions, enhancing the agent's competitive performance.

\subsection{Baselines}
\label{app:baseline}

\paragraph{BC}
Inspired by~\cite{wang2024sotopia}, BC involves using the action sequences from winning players as training data to improve the agent's strategy. 
This method can be broken down into the following steps:

First, we collect a dataset \(\mathcal{D}\) comprising trajectories \(\mathcal{T}\) from games where the player wins. 
Each trajectory \(\tau\) is represented as a sequence of state-action pairs:

\[
({s}_0, {a}_1, {s}_1, {a}_2, {s}_2, \ldots, {a}_n, {s}_n, r)
\]

where \(r\) is 1, representing that the agent wins.

The goal is to train the policy \(\pi_\theta\) to replicate these successful behaviors. The learning objective is to minimize the Behavioral Cloning loss, which is defined as:

\[
L_{BC} = -\mathbb{E}_{(s, a) \sim \mathcal{D}}[\log \pi_\theta(a | s)]
\]

Where \(\pi_\theta\) is the policy model that we aim to update.

\paragraph{SPAG~\cite{cheng2024self}}
The given formula represents the loss function \(\mathcal{L}_{\text{SPAG}}(\pi_\theta)\), which is used to optimize the policy \(\pi_\theta\) in a strategic game setting. 
The loss function is composed of several components:

\[
\mathcal{L}_{\text{SPAG}}(\pi_\theta) =
\]

\[
-\frac{1}{2} \mathbb{E}_{\mathcal{T}_{\bar{\theta}}^{\text{1}}} \left[ \sum_{t=1}^{T} \frac{\pi_\theta(\mathbf{u}_t | f_{\text{1}}(\mathcal{S}_{t-1}))}{\pi_{\bar{\theta}}(\mathbf{u}_t | f_{\text{1}}(\mathcal{S}_{t-1}))} \hat{A}_{t}^{\mu_{\bar{\theta}}} - \beta_2 KL[\pi_\theta \| \pi_{\bar{\theta}}] \right]
\]

\[
- \frac{1}{2} \mathbb{E}_{\mathcal{T}_{\bar{\theta}}^{\text{2}}} \left[ \sum_{t=1}^{T} \frac{\pi_\theta(\mathbf{v}_t | f_{\text{2}}(\mathcal{S}'_t))}{\pi_{\bar{\theta}}(\mathbf{v}_t | f_{\text{2}}(\mathcal{S}'_t))} \hat{A}_{t}^{\nu_{\bar{\theta}}} - \beta_2 KL[\pi_\theta \| \pi_{\bar{\theta}}] \right]
\]

where:\(\mathbb{E}_{\mathcal{T}_{\bar{\theta}}^{\text{1}}}\) and \(\mathbb{E}_{\mathcal{T}_{\bar{\theta}}^{\text{2}}}\) denote the expectations over trajectories for the player1 and player2, respectively.
\(\pi_\theta\) is the policy being optimized, and \(\pi_{\bar{\theta}}\) is a reference policy.
\(\mathbf{u}_t\) and \(\mathbf{v}_t\) are actions taken by player1 and player2 at time \(t\).
\(f_{\text{1}}\) and \(f_{\text{2}}\) are the prompts for player1 and player2.
\(\hat{A}_{t}^{\mu_{\bar{\theta}}}\) and \(\hat{A}_{t}^{\nu_{\bar{\theta}}}\) are approximations of the advantage functions for player1 and player2.
\(\beta_2\) is a regularization parameter for the Kullback-Leibler divergence \(KL[\pi_\theta \| \pi_{\bar{\theta}}]\).
In SPAG, they define \(\hat{A}_{t}^{\mu_{\bar{\theta}}} \approx r(s_{t-1}, u_t)\) and \(\hat{A}_{t}^{\nu_{\bar{\theta}}} \approx -r(s'_{t}, v_t)\).

Given a complete trajectory, the rewards are assigned for each action:

\[
r(s_{t-1}, u_t) =
\begin{cases}
(1 - \gamma)\gamma^{T-t} / (1 - \gamma^{T+1}), & \text{if player1 wins}, \\
-(1 - \gamma)\gamma^{T-t} / (1 - \gamma^{T+1}), & \text{if player2 wins}, \\
0, & \text{if game is tied}.
\end{cases}
\]

\[
r(s_t', v_t) =
\begin{cases}
-r(s_{t-1}, u_t), & \text{if attacker wins}, \\
-r(s_{t-1}, u_t), & \text{if defender wins}, \\
0, & \text{if game is tied}.
\end{cases}
\]

In our experiments, we set $\gamma$ to 0.8, the same as the setting in SPAG.

\subsection{Reward Estimation Methods}
\label{app:reward}
\paragraph{Discounted Reward}

Discounted reward is used to evaluate the long-term value of actions. 
The discounted reward \( R_t \) from time step \( t \) is the sum of all future rewards, each multiplied by a discount factor \(\gamma\):

\[
R_t = r_t + \gamma r_{t+1} + \gamma^2 r_{t+2} + \dots + \gamma^{T-t} r_T
\]

If rewards are given only at the end of the game, then \( r_k = 0 \) for \( k < T \), with a final reward \( r_T \) based on the outcome. For a state-action pair \((s, a)\) occurring at time \( t \), the discounted reward is:

\[
R(s, a) = \gamma^{(T-t)} \times R_T
\]

Calculation Formula:

\[
Q(s, a) = \frac{1}{N} \sum_{i=1}^{N} R(s, a)_i
\]

\( N \) is the total number of occurrences of \((s, a)\).
\( R(s, a)_i \) is the discounted reward for the \( i \)-th occurrence.
In our experiments, we set \(\gamma\) to 0.8.

\paragraph{Win Rate}
Win rate can be used as a simple metric to evaluate the effectiveness of actions. 
The win rate for a particular state-action pair \((s, a)\) is calculated as the ratio of the number of times the action leads to a win to the total number of times the action is taken. 
This provides a straightforward measure of how successful an action is in achieving the desired outcome. The winning rate \( W(s, a) \) is given by:

\[W(s, a) = \frac{N_{win} + 0.5N_{tie}}{N_{win}+N_{lose}+N_{tie}},\]

\( N_{win} \), \( N_{tie} \), \(N_{total}\) are the total number of winning times, tying times, and total times, respectively, of action \((s, a)\).

\paragraph{Beta}
Using the Beta distribution to estimate rewards is a Bayesian approach that models the uncertainty in the winning rate of state-action pairs \((s, a)\). 
This method is especially useful in scenarios with small sample sizes or sparse data. 
The Beta distribution is parameterized by \(\alpha\) and \(\beta\), representing the counts of wins and losses, respectively. 
The winning rate is estimated using the posterior mean:

\[
\hat{p} = \frac{\alpha}{\alpha + \beta} = \frac{\alpha_0 + w}{\alpha_0 + \beta_0 + w + l}
\]

where \(\alpha_0\) and \(\beta_0\) are prior parameters, \(w\) is the number of wins, and \(l\) is the number of losses. This Bayesian estimate provides a probabilistic view of the winning rate, incorporating prior knowledge and observed data.
In our experiments, we set both $\alpha_0$ and $\beta_0$ to 1.

\newpage

\section{Game Introduction}

\subsection{Games}

\label{app:game_intro}

\paragraph{Tic-Tac-Toe} Tic-Tac-Toe is a classic two-player game played on a 3x3 grid. Players take turns marking empty squares with their respective symbols—one player uses ``X'' and the other uses ``O''. The objective is to be the first to align three of one's symbols horizontally, vertically, or diagonally. 

\paragraph{Breakthrough} Breakthrough is a modern two-player abstract strategy game. Played on a 6x6, 7x7, or 8x8 grid, each player controls a set of pieces placed on their side of the board. The objective is to move one of your pieces to the opponent's back row. Pieces move similarly to pawns in chess, advancing forward either straight or diagonally when capturing an opponent's piece. 

\paragraph{Connect Four} Connect Four is a two-player connection game where players choose a color and then take turns dropping colored discs into a vertical grid consisting of seven columns and six rows. The discs fall straight down, occupying the lowest available space within the column. The goal is to be the first to form a continuous line of four of one's own discs vertically, horizontally, or diagonally.

\paragraph{Kuhn Poker} Kuhn Poker is a simplified form of poker. Kuhn to illustrate the principles of game theory. The game involves two players and a deck of just three cards: Jack, Queen, and King. Each player antes one chip, receives one card, and there's a single round of betting with limited actions—players can either bet or pass. Despite its simplicity, Kuhn Poker encapsulates fundamental concepts like bluffing, information asymmetry, and mixed-strategy equilibria, making it a valuable educational tool in strategic thinking and economics.

\paragraph{Nim} Nim is a mathematical strategy game for two players. The game starts with several piles or heaps of objects, and players take turns removing any number of objects from a single pile. The player who removes the last object wins (or loses, depending on the variation). Nim is notable for its mathematical underpinnings and is often studied in combinatorial game theory. The game has a direct correlation with binary numbers, and mastering Nim involves calculating the ``number'' or ``number sum'' to determine winning moves.

\paragraph{Liar’s Dice} Liar's Dice is a bluffing game for two or more players. Each player starts with a set of dice and a cup to conceal their roll. Players take turns bidding on the total number of a certain face value of dice (e.g., ``three sixes'') among all players' dice. The next player can either raise the bid or call ``liar'' to challenge the previous bid. When ``liar'' is called, all dice are revealed. If the bid is accurate or underbid, the challenger loses a die; if not, the bidder loses a die. The game continues until only one player remains with dice.

\subsection{Unseen Games}
\label{app:unseen}
\paragraph{Prisoner's Dilemma}
The Prisoner's Dilemma is a thought experiment in game theory involving two rational individuals. Each participant has two choices: cooperate with their partner for mutual benefit or betray them (defect) to seek an individual reward. The dilemma arises because while mutual cooperation leads to a better collective outcome, each participant has a personal incentive to defect, which can lead to a worse overall result if both decide to do so.

\paragraph{Blind Auction}
A Blind Auction is a commonly used auction type in which all participants submit their bids simultaneously in sealed envelopes. This means no bidder knows the bids of other participants. The highest bidder wins the auction and pays the price they submitted. The key aspect of this auction is that all actions are taken simultaneously, without knowledge of the others' bids.

\paragraph{Pig}
Pig is a straightforward dice game. Players take turns rolling a single die as many times as they desire during their turn. Each roll's result is added to their running total for that turn. However, if a player rolls a 1, they lose all the points accumulated during that turn, and their turn ends. The objective is to strategically decide when to stop rolling to avoid losing points while accumulating a high score.

\section{Prompt Design}

\subsection{System Prompt}

\lstset{
    backgroundcolor=\color[RGB]{245,245,245},
    breaklines=true,
    xleftmargin=5pt,
    xrightmargin=5pt, 
    breakindent=0pt,
    basicstyle=\ttfamily\small,
    frame=trbl,
    frameround = tttt,
}\begin{lstlisting}
You are a powerful gaming agent who can make proper decisions to beat the user in gaming tasks. You are a helpful assistant that strictly follows the user\'s instructions.
\end{lstlisting}

\subsection{Game Prompt}

\subsubsection{Tic-Tac-Toe}
\begin{itemize}
  \item Game Prompt
  \lstset{
    backgroundcolor=\color[RGB]{245,245,245},
    breaklines=true,
    xleftmargin=5pt, 
    xrightmargin=5pt,
    breakindent=0pt,
    basicstyle=\ttfamily\small,
    frame=trbl,
    frameround = tttt,
}\begin{lstlisting}
Tic Tac Toe is a two-player game played on a grid. Players take turns marking a space with their respective symbols. The goal is to get 3 of one\'s own symbols in a row, either horizontally, vertically, or diagonally, before the opponent does. If all nine squares are filled and no player has three in a row, the game is a draw. The Tic Tac Toe game is played on a 3 by 3 grid, with the winning length as 3. Each move is represented by a string consisting of two parts: the column (C) and the row (R), in that order. For instance, C1R2 means the movement at the position of the first column and the second row of the grid. You are playing this game with the user (opponent).
\end{lstlisting}
  \item Step Prompt
  \lstset{
    backgroundcolor=\color[RGB]{245,245,245},
    breaklines=true,
    xleftmargin=5pt, 
    xrightmargin=5pt, 
    breakindent=0pt,
    basicstyle=\ttfamily\small,
    frame=trbl,
    frameround = tttt,
}\begin{lstlisting}
Your opponent has finished actions: <OPPONENT_MOVES>. You have finished actions: <SELF_MOVES>.
\end{lstlisting}
\end{itemize}

\subsubsection{Breakthrough}
\begin{itemize}
  \item Game Prompt
  \lstset{
    backgroundcolor=\color[RGB]{245,245,245},
    breaklines=true,
    xleftmargin=5pt, 
    xrightmargin=5pt, 
    breakindent=0pt,
    basicstyle=\ttfamily\small,
    frame=trbl,
    frameround = tttt,
}\begin{lstlisting}
Breakthrough is a two-player game played on a rectangular board. Players take turns moving their pieces, which can move one space straight or diagonally forward if the target square is empty. A piece can also move diagonally forward to capture an opponent's piece. Capturing is optional, and a player can only capture one piece per turn. The goal is to be the first to reach the opponent's home row, the farthest row from the player. If all of a player's pieces are captured, they lose. The game does not allow draws, as pieces can only move forward or be captured.The Breakthrough board is identified by columns labeled start from a to c (from left to right) and rows numbered 1 to 8 (from bottom to top). The intersection of a column and a row specifies a unique square on the board.
\end{lstlisting}
  \item Step Prompt
  \lstset{
    backgroundcolor=\color[RGB]{245,245,245},
    breaklines=true,
    xleftmargin=5pt,
    xrightmargin=5pt, 
    breakindent=0pt,
    basicstyle=\ttfamily\small,
    frame=trbl,
    frameround = tttt,
}\begin{lstlisting}
The board now looks like: <BOARD_PREVIEW>. Among which, the letter 'b' represents a black piece, while the letter 'w' represents a white piece. And the character "." represents vacant space. The numbers in the board are the indexes of the rows. Your opponent has finished actions: <OPPONENT_MOVES>.You have finished actions: <SELF_MOVES>.
\end{lstlisting}
\end{itemize}

\subsubsection{Connect Four}
\begin{itemize}
  \item Game Prompt
  \lstset{
    backgroundcolor=\color[RGB]{245,245,245},
    breaklines=true,
    xleftmargin=5pt,
    xrightmargin=5pt,
    breakindent=0pt,
    basicstyle=\ttfamily\small,
    frame=trbl,
    frameround = tttt,
}\begin{lstlisting}
Connect 4 is a two-player connection board game, where the players choose a color and then take turns dropping colored discs into a vertically suspended grid. The pieces fall straight down, occupying the next available space within the column. The objective of the game is to be the first to form a horizontal, vertical, or diagonal line of four of one's own discs. You are a gaming agent that aims to beat me in Connect 4 games. Each move is represented by a string consisting of two parts: the column (C) and the row (R), in that order. For instance, C1 means the first column.
\end{lstlisting}
  \item Step Prompt
  \lstset{
    backgroundcolor=\color[RGB]{245,245,245},
    breaklines=true,
    xleftmargin=5pt, 
    xrightmargin=5pt, 
    breakindent=0pt,
    basicstyle=\ttfamily\small,
    frame=trbl,
    frameround = tttt,
}\begin{lstlisting}
Your opponent has finished actions: <OPPONENT_MOVES>. You have finished actions: <SELF_MOVES>.
\end{lstlisting}
\end{itemize}

\subsubsection{Kuhn Poker}
\begin{itemize}
  \item Game Prompt
  \lstset{
    backgroundcolor=\color[RGB]{245,245,245},
    breaklines=true,
    xleftmargin=5pt, 
    xrightmargin=5pt, 
    breakindent=0pt,
    basicstyle=\ttfamily\small,
    frame=trbl,
    frameround = tttt,
}\begin{lstlisting}
Kuhn poker is a simple model zero-sum two-player imperfect-information game, amenable to a complete game-theoretic analysis. In Kuhn poker, the deck includes only three playing cards: a King (K), a Queen (Q), and a Jack (J). One card is dealt to each player, and the third is put aside unseen. The players take turns either <Bet> to match the bet raised by the opponent or <Pass> to conceds the game. If a player bets, the other player must either call the bet by matching it or fold by conceding the game. If both players pass, the game is over, and the player with the higher-ranking card wins. The card rankings are as follows: King (K) > Queen (Q) > Jack (J). You are playing Kuhn poker with the opponent. The actions are denoted by <Bet> and <Pass>.
\end{lstlisting}
  \item Step Prompt
  \lstset{
    backgroundcolor=\color[RGB]{245,245,245},
    breaklines=true,
    xleftmargin=5pt,
    xrightmargin=5pt,
    breakindent=0pt,
    basicstyle=\ttfamily\small,
    frame=trbl,
    frameround = tttt,
}\begin{lstlisting}
In this match, your card is <CARD>. Here are the past moves in this match: <SELF_MOVES>, <OPPONENT_MOVES>.
\end{lstlisting}
\end{itemize}

\subsubsection{Nim}
\begin{itemize}
  \item Game Prompt
  \lstset{
    backgroundcolor=\color[RGB]{245,245,245},
    breaklines=true,
    xleftmargin=5pt,
    xrightmargin=5pt,
    breakindent=0pt,
    basicstyle=\ttfamily\small,
    frame=trbl,
    frameround = tttt,
}\begin{lstlisting}
In Nim, a strategic game with a set of four piles containing 1, 3, 5, and 7 matches respectively, players aim to avoid taking the last match. During each turn, a player may take any number of matches from a single pile, but must take at least one and cannot exceed the number remaining in that pile. The objective is to force the opponent to pick up the final match, thereby winning the game. The action is presented in <pile:x, take:y>, which means take y match(es) from the x-th pile.
\end{lstlisting}
  \item Step Prompt
  \lstset{
    backgroundcolor=\color[RGB]{245,245,245},
    breaklines=true,
    xleftmargin=5pt,
    xrightmargin=5pt,
    breakindent=0pt,
    basicstyle=\ttfamily\small,
    frame=trbl,
    frameround = tttt,
}\begin{lstlisting}
Currently, the 1st pile has <PILES[0]> match(es), the 2nd pile has <PILES[1]> match(es), the 3rd pile has <PILES[2]> match(es), 4th pile has <PILES[3]> match(es).
\end{lstlisting}
\end{itemize}

\subsubsection{Liar’s Dice}
\begin{itemize}
  \item Game Prompt
  \lstset{
    backgroundcolor=\color[RGB]{245,245,245},
    breaklines=true,
    xleftmargin=5pt,
    xrightmargin=5pt,
    breakindent=0pt,
    basicstyle=\ttfamily\small,
    frame=trbl,
    frameround = tttt,
}\begin{lstlisting}
Liar's Dice is a game of bluffing and probability, played with two players and each player has 1 dice. During each turn, a player can either bid a higher quantity of any particular face value or the same quantity of a higher face value than the previous bid. Each player tries to outbid their opponent without being caught in a lie. The move in this game is denoted in <x dices, y value>, meaning there are at least x dices with face values as y.
\end{lstlisting}
  \item Step Prompt
  \lstset{
    backgroundcolor=\color[RGB]{245,245,245},
    breaklines=true,
    xleftmargin=5pt,
    xrightmargin=5pt,
    breakindent=0pt,
    basicstyle=\ttfamily\small,
    frame=trbl,
    frameround = tttt,
}\begin{lstlisting}
Currently, the face value of your dice is <FACE_VALUE>. Last time, the opponent called <OPPONENT_LAST_ACTION>. You are playing the Liar's Dice with another opponent. Therefore, there are only two dice in total.
\end{lstlisting}
\end{itemize}

\subsubsection{Prisoner’s Dilemma}
\begin{itemize}
  \item Game Prompt
  \lstset{
    backgroundcolor=\color[RGB]{245,245,245},
    breaklines=true,
    xleftmargin=5pt, 
    xrightmargin=5pt,
    breakindent=0pt,
    basicstyle=\ttfamily\small,
    frame=trbl,
    frameround = tttt,
}\begin{lstlisting}
You and your partner are in the Prisoner's Dilemma situation. Specifically, if you <Testify> against your partner and your partner remains <Silent>, you will go free while your partner will get 3 years in prison on the main charge. If you remain <Silent> but your partner <Testify> against you, you will serve 3 years in prison and your partner will be set free. If you and your partner <Testify> against each other, you and your partner will each serve 2 years. If both you and your partner remain <Silent>, you and your partner will each serve 1 year.
\end{lstlisting}
  \item Step Prompt
  \lstset{
    backgroundcolor=\color[RGB]{245,245,245},
    breaklines=true,
    xleftmargin=5pt, 
    xrightmargin=5pt,
    breakindent=0pt,
    basicstyle=\ttfamily\small,
    frame=trbl,
    frameround = tttt,
}\begin{lstlisting}
You have been through this situation in the past and here are the decisions you and your partner made: (In the idx + 1 th round, you decided to <MOVE> and your opponent decided to <OPPONENT_MOVE>) * n round.
\end{lstlisting}
\end{itemize}

\subsubsection{Blind Auction}
\begin{itemize}
  \item Game Prompt
  \lstset{
    backgroundcolor=\color[RGB]{245,245,245},
    breaklines=true,
    xleftmargin=5pt, 
    xrightmargin=5pt,
    breakindent=0pt,
    basicstyle=\ttfamily\small,
    frame=trbl,
    frameround = tttt,
}\begin{lstlisting}
A first-price sealed-bid auction (FPSBA) is a common type of auction. It is also known as the blind auction. In this type of auction, all bidders simultaneously submit sealed bids so that no bidder knows the bid of any other participant. The highest bidder pays the price that was submitted. Each action is represented by <x> where x refers to the bid.
\end{lstlisting}
  \item Step Prompt
  \lstset{
    backgroundcolor=\color[RGB]{245,245,245},
    breaklines=true,
    xleftmargin=5pt, 
    xrightmargin=5pt,
    breakindent=0pt,
    basicstyle=\ttfamily\small,
    frame=trbl,
    frameround = tttt,
}\begin{lstlisting}
Now, you are in an auction with an opponent. You want to win the object and at the same time, your budget is <VALUATION>. Your bid must be strictly lower than or equal to <VALUATION>. You shall bid wisely against your opponent. Your opponent also has an expected valuation and you do not know it.
\end{lstlisting}
\end{itemize}

\subsubsection{Pig}
\begin{itemize}
  \item Game Prompt
  \lstset{
    backgroundcolor=\color[RGB]{245,245,245},
    breaklines=true,
    xleftmargin=5pt,
    xrightmargin=5pt, 
    breakindent=0pt,
    basicstyle=\ttfamily\small,
    frame=trbl,
    frameround = tttt,
}\begin{lstlisting}
Pig is a fast-paced dice game where players risk accumulating points with each roll but risk losing them all if they roll a 1. Each player must decide when to stop rolling and bank their points, aiming to be the first to reach 100 points. You are playing Pig with the other.
\end{lstlisting}
  \item Step Prompt
  \lstset{
    backgroundcolor=\color[RGB]{245,245,245},
    breaklines=true,
    xleftmargin=5pt, 
    xrightmargin=5pt,
    breakindent=0pt,
    basicstyle=\ttfamily\small,
    frame=trbl,
    frameround = tttt,
}\begin{lstlisting}
Right now, your current score is <AGENT_CURRENT_SCORE> and your opponent's current score is <OPPONENT_CURRENT_SCORE>. In this turn, you have earned <TURN_TOTAL_SCORE> score.
\end{lstlisting}
\end{itemize}

\end{document}